\documentclass[sigconf]{acmart}
\usepackage[figuresright]{rotating}
\usepackage{tabularx}
\usepackage{booktabs}
\usepackage{array}
\usepackage[utf8]{inputenc}
\AtBeginDocument{%
  \providecommand\BibTeX{{%
    \normalfont B\kern-0.5em{\scshape i\kern-0.25em b}\kern-0.8em\TeX}}}

\setcopyright{acmcopyright}
\copyrightyear{2023}
\acmYear{2023}
\acmDOI{}

\acmConference[ACM/IEEE HRI 23, workshop CONCATENATE]{}{March 13, 2023}{Stockholm, Sweden}
\acmPrice{}
\acmISBN{}

\begin{document}

\title{ On the impact of robot personalization on human-robot interaction: A review}

\author{Jinyu Yang}
\email{jinyu.yang@tum.de}
\affiliation{%
  \institution{Technical University of Munich}
  \country{Germany}
}
\author{M. Sc. Camille Vindolet}
\email{camille.vindolet@tum.de}
\affiliation{%
  \institution{Institute for Cognitive Systems}
    \institution{Technical University of Munich}
  \country{Germany}
}

\author{Dr.-Ing. Julio Rogelio Guadarrama Olvera}
\email{rogelio.guadarrama@tum.de}
\affiliation{%
  \institution{Institute for Cognitive Systems}
    \institution{Technical University of Munich}
  \country{Germany}
}

\author{Prof. Dr. Gordon Cheng}
\email{gordon@tum.de}
\affiliation{%
  \institution{Institute for Cognitive Systems}
    \institution{Technical University of Munich}
  \country{Germany}
}

\renewcommand{\shortauthors}{Yang, et al.}

\begin{abstract}
  This study reviews the impact of personalization on human-robot interaction. Firstly, the various strategies used to achieve personalization are briefly described. Secondly, the effects of personalization known to date are discussed. They are presented along with the personalized parameters, personalized features, used technology, and use case they relate to. It is observed that various positive effects have been discussed in the literature while possible negative effects seem to require further investigation.
\end{abstract}




\received{27 January 2023}

\maketitle

\section{Introduction}
Nowadays, robots are expected to evolve alongside humans and to interact more closely with them. In particular, socially assistive robots (SAR) should be capable of social interactions and are presented as a potential solution for the growing need for assistance in domains such as healthcare, education, and elderly care \cite{rouaix_affective_2017}. They should, for example, perform various household tasks, provide companionship to the elderly, assist children in learning \cite{tanev_implementation_2022}, deliver meals \cite{dino_delivering_2019}, or guide people in tours of shopping \cite{nithya_priya_autonomous_2021} or museums. 

As robots are expected to become increasingly integrated into people's daily lives, it seems important to create more natural and personalized interactions between humans and robots.
Indeed, they should become companions that provide both mental and physical support. This type of interactions require an emotional intelligence from the robot side with a capacity to adapt to the user's state \cite{duraes_characterize_2018}. Thus, personalization is one of the current field of research in HRI and more specifically in SAR. It aims to enable robots to understand human needs and preferences and to correctly adapt to them.  For example, personalized robots should recognize and respond accordingly to the users social interactions, nonverbal behaviors, or culture. Personalization is said to improve the user experience, encourage social acceptance of the robot, and lead to more effective social interactions by adapting the behavior, appearance and interaction patterns of robots to the individuals needs and preferences.

Many strategies for achieving personalization have been investigated. Mehdi Hellou et al. (2021) \cite{hellou_personalization_2021} presented a review of the employed methods, to date, to create robot personalization. 

Personalization can influence differently the user perception of the quality of the interaction and of the system trustworthiness. It can help improving the match between the user expectations and the robot behavior. This paper reviews the findings on the impact of personalization in HRI, in order to acquire a more comprehensive understanding of how personalization affects the interaction experience and provided service that better fulfills the requirements of the user. Examples of current personalization strategies are provided in section 2. Personalizing a robot for different scenarios and work contexts is possible in a variety of ways. The consequences of various personalization are studied in section 3. Section 4 discusses possible directions for further research into the effects of personalization.

\section{Personalized Robot Strategies and use cases}

\begin{table*}[thp]
    \centering
    \begin{tabularx}{\textwidth} {  >{\raggedright\arraybackslash}X  >{\raggedright\arraybackslash}X  >{\raggedright\arraybackslash}X  >{\raggedright\arraybackslash}X >{\raggedright\arraybackslash}X >{\raggedright\arraybackslash}X} 
    \toprule
    &  \textbf{Analysed Parameters} 
    &  \textbf{Personalized Features}
    &  \textbf{Technology, Hardware and Sensors}
    &  \textbf{Use Cases/ Purpose}
    &  \textbf{The Effects of personalization} \\
    \midrule
    M. K. Lee et al. (2012) \cite{lee_personalization_2012}
    & Users' snack selection, service-patterns and previous behaviour of the robot
    & Social interactions and small talk topics by robot
    & Snack ordering website, Snackbot, SICK LIDAR, etc.
    & Delivering snacks 
    & Reinforcement of people's rapport, cooperation and engagement with robot
    \\
    \midrule
    D. Portugal et al. (2015) \cite{portugal_socialrobot_2015}
    & Users' emotional state and personal preferences
    & Robot's service: movement, navigation, emotional companionship, etc.
    & Service-Oriented Architecture (SOA), service-oriented architecture, etc.
    & Home care of the elder
    & Not mentioned
    \\
    \midrule
    G. Gordon et al. (2016) \cite{gordon_affective_2016}
    & Emotional and game state 
    & Game-related responses (instructions, prompts) and emotional responses
    & Tega robot platform, Robot Operating System (ROS), etc.
    & Regulating the student's emotional state while providing them with second language tutoring 
    & Significant increase in children’s engagement and valence
    \\
    \midrule
    C. E. Clabaugh (2017) \cite{clabaugh_interactive_2017}
    & Frequency of elicitation (asking learning-sensitive questions)
    & Robotic tutor’s elicitation of learning-sensitive information
    & Not mentioned
    & Robot tutor for personalized education
    & Improved stimulation and increased trust towards the robot 
    \\
    \midrule
    N. Churamani et al. (2017) \cite{churamani_impact_2017}
    & User utterance and personal information
    & Utterance by robot
    & NICo, Humanoidly Speaking Scenario, Face and Speech Recognition, etc.
    & Teaching the robot to recognize different objects
    & Improved impressions of the robot for the users
    \\
    \midrule
    D. Leyzberg et al. (2018) \cite{leyzberg_effect_2018}
    & Participants' level of skill acquisition 
    & Ordering of the interchangeable lesson chapters
    & The extended Bayesian Knowledge Tracing (BKT) family of student models with camera and microphone
    & Rearrange curriculum while completing an English Language Learning educational task
    & Increased learning gains
    \\
    \midrule
    C. Clabaugh et al. (2019) \cite{clabaugh_long-term_2019}
    & Children's behaviour during the game (Whether mistakes were made, help requested, etc.)
    & Level of challenge and feedback
    & Tega robot platform, electrodermal activity sensors, etc.
    & Social and educational development of children with ASD
    & Improved learning and evidence of longer-term engagement
    \\
    \midrule
    H. W. Park et al. (2019) \cite{park_model-free_2019}
    & Users’ behavior and affective arousal state
    & Lexical and syntactic complexity in storytelling
    & Tega robot platform, electrodermal activity sensors, etc.
    & Learning companion
    & Increase of Children's engagement, improved learning, increased use of the target words, etc.
    \\
    \midrule
    S. Schneider and Franz Kummert (2021) \cite{schneider_comparing_2021}
    & Level of automation
    & Robot's service: propose different activities
    & Database of different exercises, session controller, 3D depth sensor, etc.
    & Assisting the user in exercising
    & Higher trust and perceived as more competent
    \\
    \midrule
    M. E. U. Ligthart et al. (2022) \cite{ligthart_memory-based_2022}
    & Children's interest, preferences and choices 
    & Serial narrative dialog
    & Google's dialog flow, onboard microphone, Redis database, etc.
    & Communicating with children in dialogue
    & Longer interest for the robot, better engagement
    \\
    \bottomrule     
    \end{tabularx}
    \caption{Summary of the papers analyzed in this review.}
    \label{tab:tablep1}
\end{table*}

As discussed in the previous section, there are various use cases of SAR, such as visiting of and caring for the elderly, helping disabled people with rehabilitation training, assisting and instructing children in their studies, among others. As a result, there is not a single solution for personalizing a robot. The robot's personalization and the researchers' design can vary from simple methods, such as choosing between several pre-defined dialogue or actions the one that fit the best the user needs and preferences, to more complex algorithms and technology that, based on multisensory integration and complex models, generate appropriate robot behaviors. The examples studied in this article are listed in Table \ref{tab:tablep1}. The table presents in order of the columns, the article, the experiment parameters taken into account for the personalization, the personalized robot features, the encountered technology, the use case and purpose of the personalization and its effects.\\

\begin{figure}
    \centering
    \includegraphics[scale=0.45]{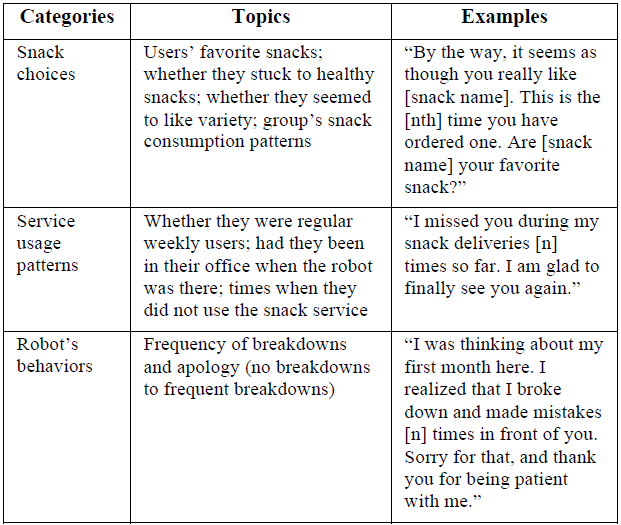}
    \caption{An overview of the personalized topics proposed by Min Kyung Lee et al. \cite{lee_personalization_2012}}
    \label{fig:1}
\end{figure}

Personalization can be, for example, pre-programming the robot's speech or actions based on user data. To investigate how to build and maintain rapport between robots and people, Min Kyung Lee et al. (2012) \cite{lee_personalization_2012} designed a snack delivering robot and created a personalization strategy that involved developing pre-defined discussions and responses, later selected based on the user's prior interactions with the robot and with its delivery service. The robot will present multiple discussions to the user depending on the user's snack selection pattern, and service usage pattern, and on the robot's past behavior (as indicated in figure \ref{fig:1}). Based on a record of defects and faults maintained in the service database, the robot also apologizes for any errors that have happened.

More recently, researchers have developed complex algorithms and techniques to create personalized robot strategies. These algorithms frequently require a combination of data, such as how the participant interacts with the robot, the responses given to the robot's activities (e.g., facial expressions, etc.), and the history of previous interactions. \\

David Portugal et al. (2015) \cite{portugal_socialrobot_2015} provided a summary of the techniques developed in the SocialRobot project. In this project, a mobile robot platform was created in conjunction with virtual social care technologies in order to enhance elder's quality of life and fulfill their individual needs. They created a model based on service-oriented architecture (SOA) and used social care community network to manage and coordinate user profiles for proactive and individualized care. The database could dynamically be adjusted in order to update user preferences.

Goren Gordon et al. \cite{gordon_affective_2016} created an integrated system of affective policy learning and autonomous social robots to investigate how effectively personalized robots regulate students' affective states in one-on-one tutorials. Various states were specified as parameters in the affective policy: the child's valence, engagement, as well as information about whether they correctly answered the questions and if they had previously interacted with the robot. A state-action reward state-action (SARSA) algorithm and an epsilon-greedy algorithm are used to determine the robot's action states. 

Hae Won Park et al. \cite{park_model-free_2019} studied how robots can assist young learners in optimizing their learning results. To manage the selection of robot behaviors, they created an affective personalization policy. The user's behavior (their question answering behavior during
the robot’s story narration), their affective arousal state (their facial muscle activation illustrating their expressiveness and arousal level), the robot's actions (the lexical and syntactic complexity of a given sentence in storytelling), the rewards the robot earned for its activities (the new lexical and syntax learning), and the changes in the user's state were organized into different episodes that were trained by Q-Learning, resulting in a personalized policy for each user.

Caitlyn Clabaugh et al. \cite{clabaugh_long-term_2019} developed a hierarchical framework for Human Robot Learning to manage the behavior of personalized robots in their study of assistive robots serving children with autism spectrum disorders (ASD). In this framework, the robot's interventions were subdivided into deﬁned state-action sub-spaces, with five controllers each responsible for a subset of actions and describing abstract action categories such as disclosure, promise, instruction, feedback, and inquiry. These controllers were combined with algorithms including Q-Learning to ensure that the robot's feedback and instruction challenge levels could be personalized to each child's particular learning style.

\section{The impact of personalization}
Personalization, as discussed in the preceding section, has become an increasingly important component of HRI. Additionally, it has created new possibilities for SAR by enabling more efficient and satisfying user experiences. However, it is necessary to examine the potential impacts of personalization. This section examines both positive and negative effects of personalization including potential long-term ones.

\subsection{Positive Impact}
Caitlyn E. Clabaugh et al. (2017) \cite{clabaugh_interactive_2017} analysed the elicitation of learning-sensitive information by a robotic tutor that would be utilised by an interactive machine learning approach for personalized education. They focused on the relationship between the frequency of elicitation and human learners' impressions of tutorial interactions. The study showed that more frequent elicitation of learning-sensitive information by the robot tutor led to a significant increase in participants' interaction with the robot tutor, and that the robot tutor was perceived to be more interactive and affectionate, and was better trusted.

A study on the impact of various SAR automation levels on user perception of the system in an exercise scenario was presented by Sebastian Schneider et al. (2021) \cite{schneider_comparing_2021}. According to the study, adaptive robots (where an adaptive process can be initiated automatically by the robot, such as when the robot chooses exercises for the user) are perceived as warmer and more competent than adaptable robots (where the user adapts the system themselves, such as when the user selects their own exercises). Users trusted the adaptive robot substantially more than the adaptable robot. Moreover, users who were exposed to an adaptive robot more frequently volunteered to repeat the task.

According to Daniel Leyzberg et al. \cite{leyzberg_effect_2018} evaluated the advantages personalized social robots can offer in a learning environment. They developed a strategy to organize lessons based on a model for evaluating students' skill competency, with the content of the lessons related to educational activities for native Spanish-speaking students learning English. The study's findings demonstrated that, assuming that all students had roughly the same amount of prior knowledge, the group receiving the personalized program learned more than the group receiving the non-personalized program, which was represented in a large increase in learning gains. When compared to learners who received non-personalized education, those who took personalized courses produced a higher percentage of correct answers on tests.

Additionally,  Mike E.U. Ligthart et al. \cite{ligthart_memory-based_2022}, a memory-based personalization strategy was developed and deployed in combination with a continuous narrative interaction structure to address demands and keep children from losing interest in the robot once the novelty wore off. The research indicates that in the classic control condition, as the number of interaction sessions increased, children's interest in continuing the contact began to decrease over time. However in the personalization strategy-using condition, children's interest in continuing the interaction remained strong. Moreover, the children's proximity to the robot was higher after multiple interaction sessions when the personalization strategy was deployed, and the strategy encouraged a closer relationship between the child and the robot. Positive social cues were elicited more often.

The benefits of personalization have been extensively studied. Many of these effects are connected to specific use cases and implementation contexts. They have shown to improve the user experience and generate positive emotions.

\subsection{Possible negative Impacts}
Personalization has recently been a popular study topic. Numerous research focused on the benefits of personalization, such as increasing the user engagement. Those studies usually address the effects of personalization with regard to some hypothesis and discuss whether or not personalization has a positive impact on the features they want to study or no impact at all. For example, Mike E.U. Ligthart et al. \cite{ligthart_memory-based_2022} notes that personalization does not have a substantial impact on the need for familiarity and similarity, in addition to the obvious positive effects. However, little research has thoroughly investigated the negative impact of personalization. As a result, this chapter presents the existing research on the topic, as well as discusses other features of personalization that are likely to have a negative influence. \\

Research and discussions on the privacy policies provided by various consumer robot companies were performed by Anna Chatzimichali et al. (2021) \cite{chatzimichali_toward_2021}. According to the study, the structure, context, and clarity of privacy provisions differ considerably between companies. This highlights the topic of how personalization may have a negative impact on personal privacy, user rights, and consumer rights in a context where consumer robotics businesses are not now uniformly considering privacy issues. As personalization depends on private data such as users' individual preferences, behaviors, and interaction history, the question of how to handle this information so that privacy and user rights meet users' requirements needs to be addressed. This is also mentioned in Jim Torresen's work (2021) \cite{torresen_ethical_2021}. He examines how sensor data collection should be limited in uncommon and urgent situations and establishes the argument that privacy requires a trade-off on this. In the case of robot-assisted care for the elderly, for example, the robot is only permitted to transmit data about the elderly in order to alert caretakers, etc., if an abnormal incident occurs.

The research by Nikhil Churamani et al. \cite{churamani_impact_2017} presents an additional hypothesis regarding negative impacts. They investigated robots' social acceptability, perceived intelligence, and likability in HRI scenarios. Since more complicated interactions may raise certain user concerns, the study indicated that the personalized interactive system scored worse in terms of safety. Furthermore, because the personalized robot was equipped with speech technology that was insufficient to support its capacity to reliably recognize the user's words in a noisy setting, participants did not recommend the system for usage in the real world. This raises the possibility that personalization might have a negative impact, since a personalized robot may be unable to deal with the different disturbances and emergencies that come in real-life work environments before the technology is developed, leading people to distrust the robot and choose a more general and safe robot. In addition to human trust in the robot, whether the robot models the human as trustworthy can have an effect on the process and experience of HRI. According to research by Dylan P. Losey et al. \cite{losey_robots_2019}, modeling the human as a trusting agent can increase HRI effectiveness, however trusting humans in this scenario are rewarded less than humans treated as rational.

In addition to the user experience, considering the robot's ethical and moral feasibility is essential. For instance, the system needs to have the ability to guarantee the user's right as well as the fairness and verifiability of the decision-making process \cite{torresen_ethical_2021}. Kathrin Pollmann et al. \cite{pollmann_entertainment_2023} also discusses the necessity of maintaining an equilibrium between ethical considerations and user experience design concepts in their research. In order to avoid manipulating the user and infringing on their autonomy, it is crucial to strive to avoid personalization methods that could result in affective mechanisms (ie. "guilt trip, emotional blackmail, erotic attraction,etc." \cite{pollmann_entertainment_2023}).

Research on the potential negatives effects of personalization is currently lacking. The impact of personalization on security of user data privacy, social acceptability, trust and ethical considerations of the robots should be further researched in the future.

\section{Discussion and Conclusion}

Whereas much research has shown the positive advantages of personalization in various ways and for different use cases, such as creating positive emotional impulses for users or improving tutoring efficiency, the negative implications have not yet been properly investigated. In the previous section, some of the potential negative features of personalization were briefly discussed: privacy concerns, trust between human and robot, and consideration for ethics. Furthermore, there are a variety of issues that must be addressed in a future study on personalization.

Cong Li (2016) noted the difference between actual personalization and perceived personalization in his work on personalization \cite{li_when_2016}. He tested the personalization effect through the message sender's actual personalization process and identified that it may have misleading effects. According to his research, a real personalization process does not always result in beneficial results, and participants only reacted positively to messages that they thought were more personalized, even though these messages may not have been produced through the personalization process. He suggested that there may be certain conditions under which a message is interpreted as personalized or non-personalized. Unquestionably, when a message contains incorrect information, the receiver of the message and the message are mismatched. Such a message will cause the receiver to perceive it as non-personalized, resulting in a negative attitude.

Another concern is human acceptance or trust towards robots. This is a complex, multifaceted problem. The previously indicated negative human evaluations due to the robot's lack of competency can be linked to trust and acceptance of the robot. Leila Takayama et al. (2009) \cite{takayama_influences_2009} investigated factors influencing proxemic behaviors in HRI. It was discovered that the size of the participant's personal space when interacting with a robot was related to a number of parameters. By examining the comfortable distance between the human and the robot during HRI, S. M. Bhagya et al. (2019) \cite{bhagya_exploratory_2019} studied human affinity preferences for service robots. They discovered that happy and sad facial expressions on the robot had closer proxemics, whereas anger and disgust had the opposite effect. The comfort distance of HRI is alternatively impacted by the robot's internal sounds and human-like appearance. Additionally, Dylan P. Losey et al. (2019) \cite{losey_robots_2019} discovered that the user experience is impacted by the human trustworthiness or untrustworthiness settings when the robot models people. Generally, in addition to the influence of personalization, variables such as perceived personalization and people's acceptance of and trust in robots need to be considered and further investigated. \\

In conclusion, this review on the impact of robot personalization on HRI presented various personalization strategies, from the most simple ones to some more recent complex multimodal ones. It highlighted some strong personalization benefits that have been studied in research in the past few years. Finally this review discussed some possible negative impacts of personalization and informed on the lack of research on this topic. Overall, the effects of personalization seem to still required further investigation.\\

\bibliographystyle{ACM-Reference-Format}
\bibliography{sample-lualatex}

\appendix

\end{document}